\documentclass[letterpaper, 10 pt, conference]{ieeeconf}  

\IEEEoverridecommandlockouts                              

\usepackage{graphicx}
\usepackage{bm}

\usepackage{amsmath} 
\usepackage{amssymb}  
\usepackage{gensymb}  
\usepackage{cite}
\usepackage{amsmath,amssymb,amsfonts}
\usepackage{algorithm}
\usepackage{algpseudocode}
\usepackage{textcomp}
\usepackage{xcolor}
\usepackage{url}
\usepackage{diagbox}
\usepackage{slashbox}

\usepackage{physics}
\usepackage{amsmath}
\usepackage{tikz}
\usepackage{mathdots}
\usepackage{yhmath}
\usepackage{cancel}
\usepackage{color}
\usepackage{siunitx}
\usepackage{array}
\usepackage{multirow}
\usepackage{amssymb}
\usepackage{gensymb}
\usepackage{tabularx}
\usepackage{booktabs}
\usetikzlibrary{fadings}
\usetikzlibrary{patterns}
\usetikzlibrary{shadows.blur}
\usetikzlibrary{shapes}

\def\etal{{\it et al.}}

\title{\LARGE \bf
An analytical diabolo model for robotic learning and control
}

\author{Felix von Drigalski$^{\ast1}$, Devwrat Joshi$^{\ast\dagger1,3}$, Takayuki Murooka$^{\dagger1,4}$,\\
Kazutoshi Tanaka$^{1}$, Masashi Hamaya$^{1}$ and Yoshihisa Ijiri$^{1,2}$ 

\thanks{\textsuperscript{$\ast$}Authors contributed equally.}
\thanks{\textsuperscript{{$\dagger$}} Work done at OMRON SINIC X Corp. as part of an internship.}
\thanks{$^{1}$Felix von Drigalski, Masashi Hamaya, Kazutoshi Tanaka and Yoshihisa Ijiri are with OMRON SINIC X Corporation, Hongo 5-24-5, Bunkyo-ku, Tokyo, Japan 
        {\tt\small \{ f.drigalski,  masashi.hamaya, kazutoshi.tanaka, yoshihisa.ijiri \}@sinicx.com}}%
\thanks{$^{2}$Yoshihisa Ijiri is with OMRON Corporation, Konan 2-3-13, Minato-ku, Tokyo, Japan 
        {\tt\small yoshihisa.ijiri @omron.com}}%
\thanks{$^{3}$Devwrat Joshi is with the University of Osaka, Hosoda Laboratory
        {\tt\small devwratjoshi1234@gmail.com}}%
\thanks{$^{4}$Takayuki Murooka is with the University of Tokyo, JSK Laboratory
        {\tt\small takayuki5168@gmail.com}}
}

\begin{document}

\maketitle
\thispagestyle{empty}
\pagestyle{empty}


\begin{abstract}
In this paper, we present a diabolo model that can be used for training agents in simulation to play diabolo, as well as running it on a real dual robot arm system.
We first derive an analytical model of the diabolo-string system and compare its accuracy using data recorded via motion capture, which we release as a public dataset of skilled play with diabolos of different dynamics.
We show that our model outperforms a deep-learning-based predictor, both in terms of precision and physically consistent behavior.
Next, we describe a method based on optimal control to generate robot trajectories that produce the desired diabolo trajectory, as well as a system to transform higher-level actions into robot motions.
Finally, we test our method on a real robot system by playing the diabolo, and throwing it to and catching it from a human player.
\end{abstract}


\section{Introduction}

Movement arts such as dance, acrobatics and juggling have been enchanting audiences for millennia, and have recently become a popular challenge for robotics researchers, both for entertainment and teaching purposes~\cite{entertainment1, entertainment2, entertainment3}, and to push the limits of current technology.
Juggling, in particular, has seen increased interest\cite{ploeger2020juggling, catch-two-robots, Kizaki2012} starting from the 1990's, as robotic actuators become more capable of executing these high-acceleration tasks.
It is a non-prehensile manipulation task, which is one of the most challenging types of manipulation tasks~\cite{RuggieroRAL2018}.

The diabolo is a juggling prop that has received very little attention, likely due to its complex and unstable dynamics, which are difficult to model and formulate control laws for. 
However, following the recent advances in machine learning and papers such as~\cite{murooka-diabolo}, a diabolo-playing robot appears increasingly tractable.

While investigating the diabolo as a learning problem, we realized that it offers a number of favorable properties for robot learning.
For example, mistakes during play rarely cause sudden failure, but only a change in the diabolo's orientation, leading to dense rewards in reinforcement learning settings compared to the other juggling tasks.
Recent platforms such as OpenAI Gym~\cite{openai} offers learning environments, whose dynamics has continuous state and action spaces close to the real-robot systems.
Unlike the platform provides few tasks on each environment (e.g. the agent learns to only move forward in Ant and Hopper environments~\cite{openai}), the diabolo environment can offer more various and agile tasks in the same environment, and pose new and interesting problems for the fields of continual learning or transfer learning~\cite{FranccoisFTM2018}.
For example, we can start from learning to accelerate the diabolo, then how to stabilize it, and then gradually increase the difficulty from easier tricks, such as moving in figure patterns, to more complex ones. 
We can also apply the learned agents to other diabolo environments, which have different dynamics depending on the string length and diabolo geometry and weight. 
Considering this, the diabolo task could be a new and suitable environment for testing learning algorithms.

\begin{figure}[t!]
    \centering
    \includegraphics[width=0.85\columnwidth]{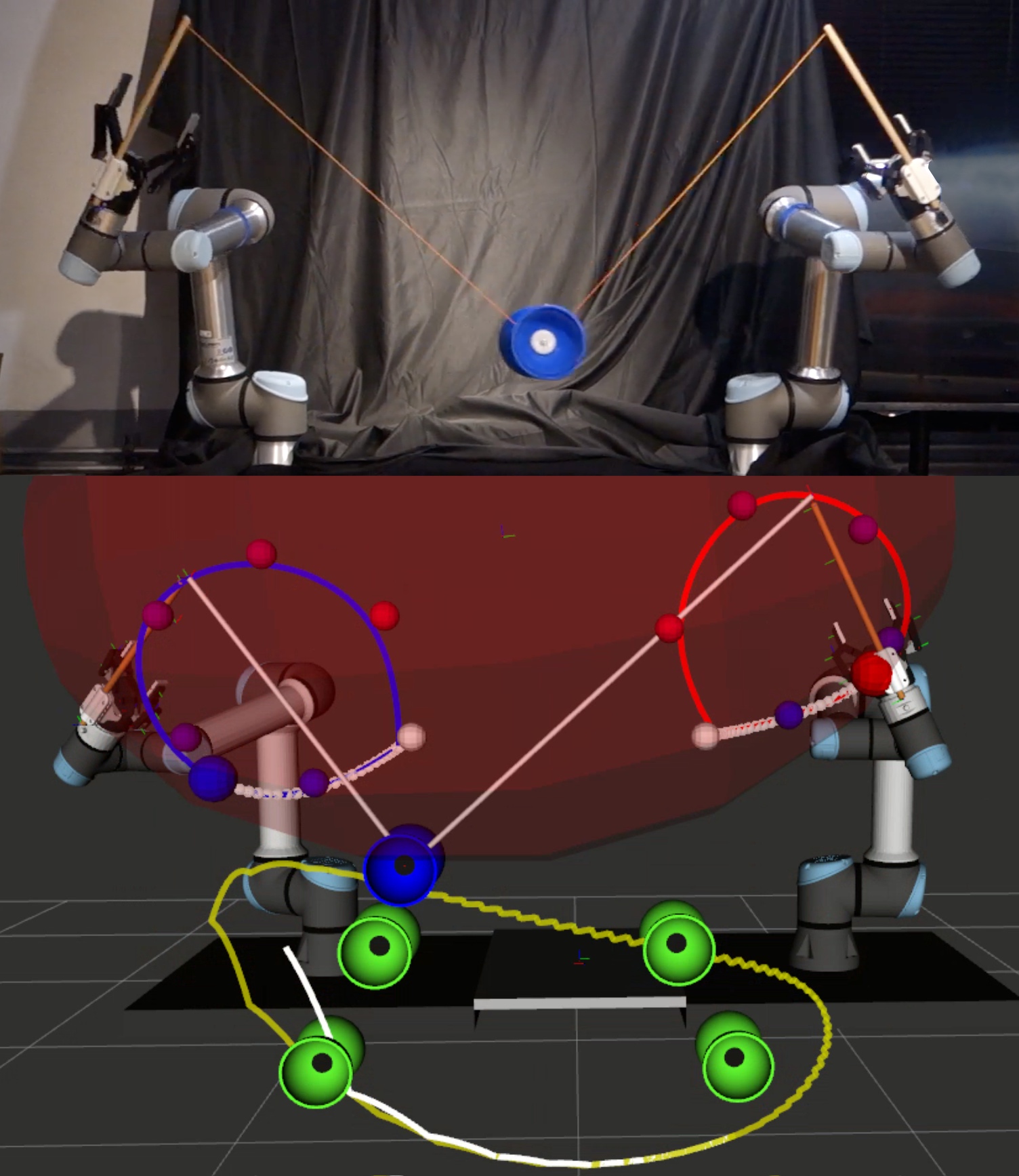}
    \caption{\textbf{Top:} Robots playing diabolo (or ``kongzhu'' or ``Chinese yo-yo''), a juggling prop that is accelerated by a string attached to the two sticks which the player manipulates to perform tricks.\\
    \textbf{Bottom:} Internal representation of our proposed baseline method. \textit{Yellow:} Predicted diabolo trajectory. \textit{Green:} Goal states. \textit{White:} Prediction until the start of the next trajectory. \textit{Red/blue:} Stick trajectories with via points. \textit{Red transparent:} Spheroid used to calculate the string effects.}
    \label{fig:intro}
\end{figure}

As setting up and supervising two real robot arms to learn diabolo playing is expensive and fast-moving robot arms can be dangerous, it would be desirable to train agents in a simulated environment and thus make learning more practical.
However, there are currently no environments that describe the diabolo's behavior.
Consequently, we propose an analytical model to predict the motion of the diabolo in simulation which can be used in a learning environment, as well as a model-based control approach which should serve as a baseline solution to the problem, to which learning agents can be compared.

The contributions of this paper are as follows:
\begin{itemize}
    \item A diabolo model generating realistic diabolo trajectories 
    \item An optimal-control-based method that determines stick trajectories which produce the desired diabolo state
    \item A robot system that plays diabolo and can react by adjusting the stick trajectories based on the system state
    \item A dataset of high-precision diabolo and stick positions during different diabolo play situations to train learning agents and evaluate our proposed model
\end{itemize}

We release the diabolo model and code along with a wrapper for the open-source simulator Gazebo~\cite{gazebo}, so it can be used as a plugin with existing robot systems or stand-alone.

The remainder of this paper is structured as follows. 
We describe related works in Section~\ref{sec:relatedwork}.
The proposed method is described in detail in Section~ \ref{sec:method}, and the experimental validation is shown in Section~\ref{sec:experiment}. 
We discuss our results in Section~\ref{sec:discussion} and conclude in Section~\ref{sec:conclusion}.


\section{Related Work}\label{sec:relatedwork}

\subsection{Juggling with robots}

Many juggling disciplines have been tackled in research to varying degrees of success and completeness. 
The following studies focus mostly on the most fundamental techniques for each type of juggling, such as throwing and catching a ball, rolling off and returning a yo-yo, or keeping a devilstick in the air.

The history of robot juggling has been well described by Ploeger \etal~\cite{ploeger2020juggling}, who most recently used reinforcement learning to train a robot arm to juggle two balls using a cup, using a model that outputs via-points and their speed, which is similar to how we represent diabolo stick trajectories.
Ball catching and throwing have been realized with two robots~\cite{catch-two-robots}, and one-handed catching and throwing balls from and to human participants~\cite{disney:catch-ball}.
Kizaki \etal{} maintained a two-ball cascade with a high-speed robot arm and an actuated end effector~\cite{Kizaki2012}.
Reist \etal{} designed a paddle with a one-degree-of-freedom actuator that can keep a bouncing ball centered due to its concave shape, and extended this ``Blind Juggler'' into various arrangements~\cite{Reist2012}.

Devilstick manipulation with a mechanical aid and 3-ball juggling was also realized by hydraulic robotic arms by Schaal \etal~\cite{devil-stick2, devil-stick}.
Flowerstick (similar to devilstick) manipulation was also addressed with visual feedback control using a high-speed vision system~\cite{flower-stick}. 
Contact juggling was simplified to a disk-on-disk system and demonstrated with closed loop~\cite{contact-juggling2} or feedback stabilization control in~\cite{contact-juggling, contact-juggling3}.
Periodic control to realize a yo-yo's up-and-down motion has been demonstrated with a robotic hand~\cite{yo-yo3, yo-yo2, yo-yo}.

\subsection{Diabolo modeling}

The interactions between the string and the diabolo are highly non-linear---we have found no analytical model in the literature, and very few systematic investigations of diabolo behavior and control.
Sharpe  \etal~\cite{SharpeDiaboloAnalysis2008} used a strobe light to compare the effect of different diabolo acceleration techniques, reporting maximum rotation speeds of 5700~rpm.
Guebara \etal~\cite{DiaboloSkillAnalysis2018} collected diabolo play data with motion capture setups similar to ours.

Murooka \etal~\cite{murooka-diabolo} proposed a deep-learning-based diabolo stabilization control method which was demonstrated with a mobile humanoid robot.
However, their method assumes that the diabolo does not move, and the sticks always perform the same motion.
Applying the method to other diabolo motions and tricks tasks would require retraining and collecting appropriate training data, which can be a challenging and uncertain process.  
In our work, we formulate an analytical model that can be used for arbitrary diabolo manipulations and adjusted by setting meaningful physical parameters.


\section{Method}\label{sec:method}

Our method consists of three main parts:

\begin{enumerate}
    \item A \textbf{predictor} model estimating the diabolo's motion as a function of the stick tips' motion (``stick trajectory'')
    \item A ``\textbf{player}'' that uses the predictor to find a stick trajectory that results in a target diabolo position and velocity
    \item A \textbf{robot control} module that monitors the diabolo state and blends new trajectories
\end{enumerate}

\begin{figure}[t]
    \centering
    \includegraphics[width=0.99\columnwidth]{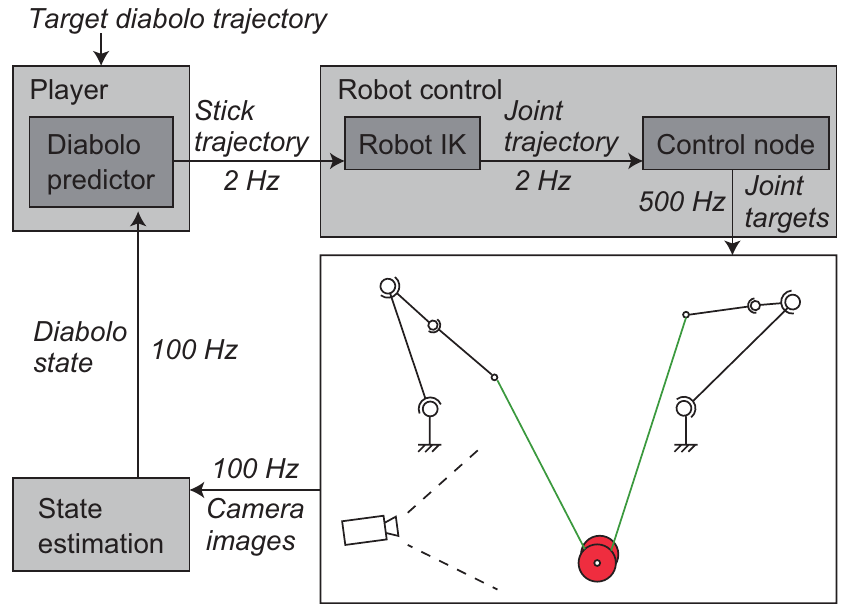}
    \caption{The system controlling the diabolo play. The player node receives high-level goals such as ``accelerate with technique A'', ``accelerate with technique B'', ``throw upwards'', ``throw to the side''.}
    \label{fig:system-layout}
\end{figure}

The system layout is shown in Fig.~\ref{fig:system-layout}.

\subsection{Diabolo Predictor}

The goal for the predictor is to estimate the state of the diabolo at time $t$, given the diabolo state at time $t-1$ and the stick positions at time $t$.
This predictor can then be used A) to simulate the diabolo behavior in a physical simulator or learning environment, or B) to evaluate (or "imagine") the effect of stick trajectories in the player module described in the next section.
To construct an analytical model of the diabolo and string system, we simplify the complex tribological interactions between the string and the diabolo by using an auxiliary spheroid (ellipsoid of rotation).

\begin{figure}[t]
    \centering
    \includegraphics[width=0.99\columnwidth]{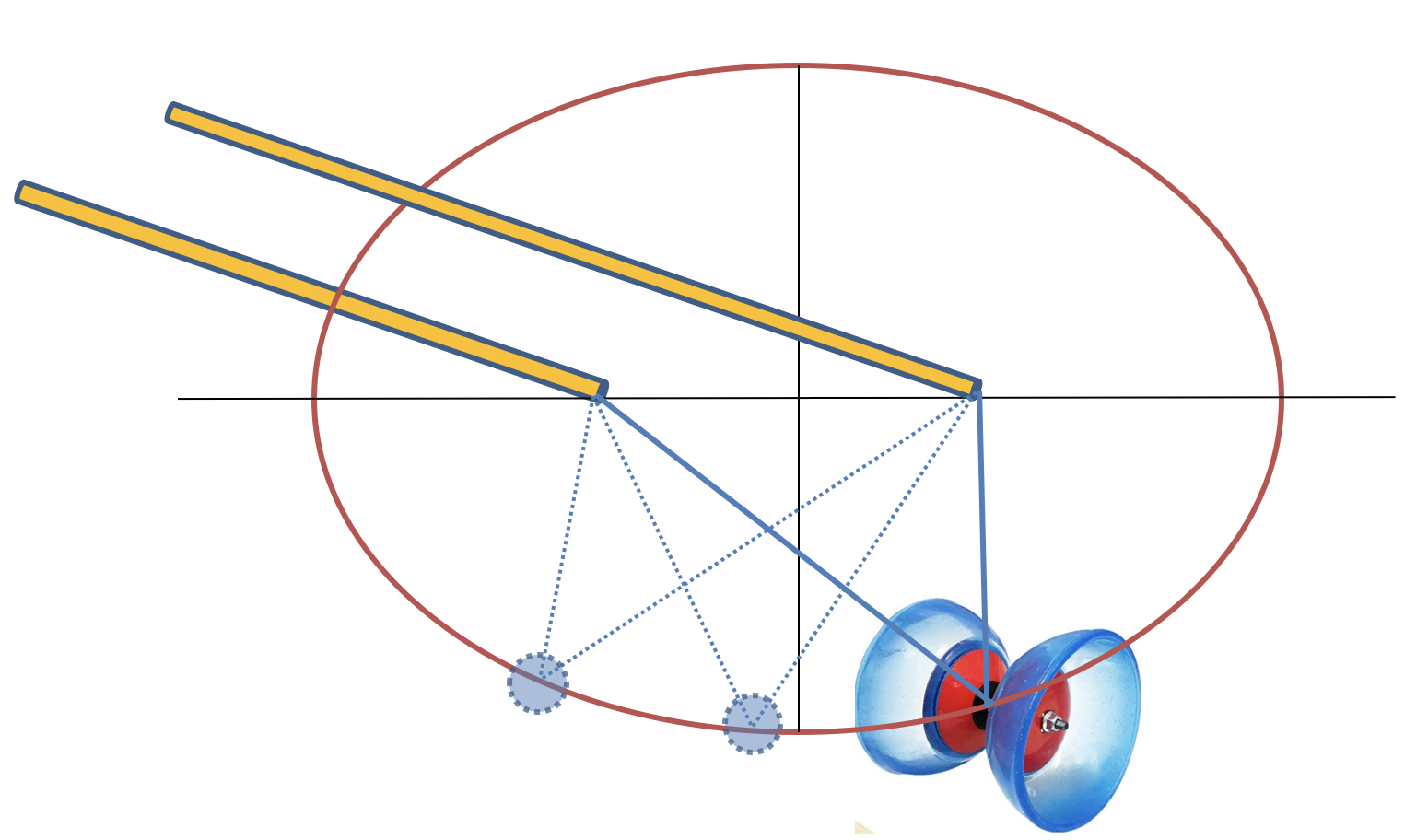}
    \caption{The ellipse which defines the auxiliary spheroid used to evaluate the diabolo position during play. The ellipse's focal points are at the stick tips. The resulting spheroid describes the points the diabolo can reach when the string is taut.}
    \label{fig:ellipse}
\end{figure}

A spheroid can be defined by two focal points, where for all the points on the spheroid, the sum of the distances to the focal points is equal.
As shown in Fig.~\ref{fig:ellipse}, this conveniently represents the points that can be reached by the diabolo when the string between the two sticks is taut.
We construct this spheroid by rotating the 2D ellipse about its major axis, with the stick tips as its focal points. 
Using the local coordinate system centered between the tips of the sticks, the semi-major and semi-minor axis $b_\mathrm{ellipse}$ and $a_\mathrm{ellipse}$ of this ellipse are calculated as follows from the stick positions $\vec{x}_\mathrm{left}$ and $\vec{x}_\mathrm{right}$ and the length of string $l_\mathrm{string}$ between the sticks as follows.

\begin{subequations}
  \label{equation:ellipse}
  \begin{align}
    a_\mathrm{ellipse} &= l_\mathrm{string}/2 \\
    b_\mathrm{ellipse} &= \sqrt{a_\mathrm{ellipse}^2 - \frac{\left \| \vec{x}_\mathrm{left} - \vec{x}_\mathrm{right} \right \|}{2}}
  \end{align}
\end{subequations}

\begin{figure}[t]
    \centering
    \includegraphics[width=0.99\columnwidth]{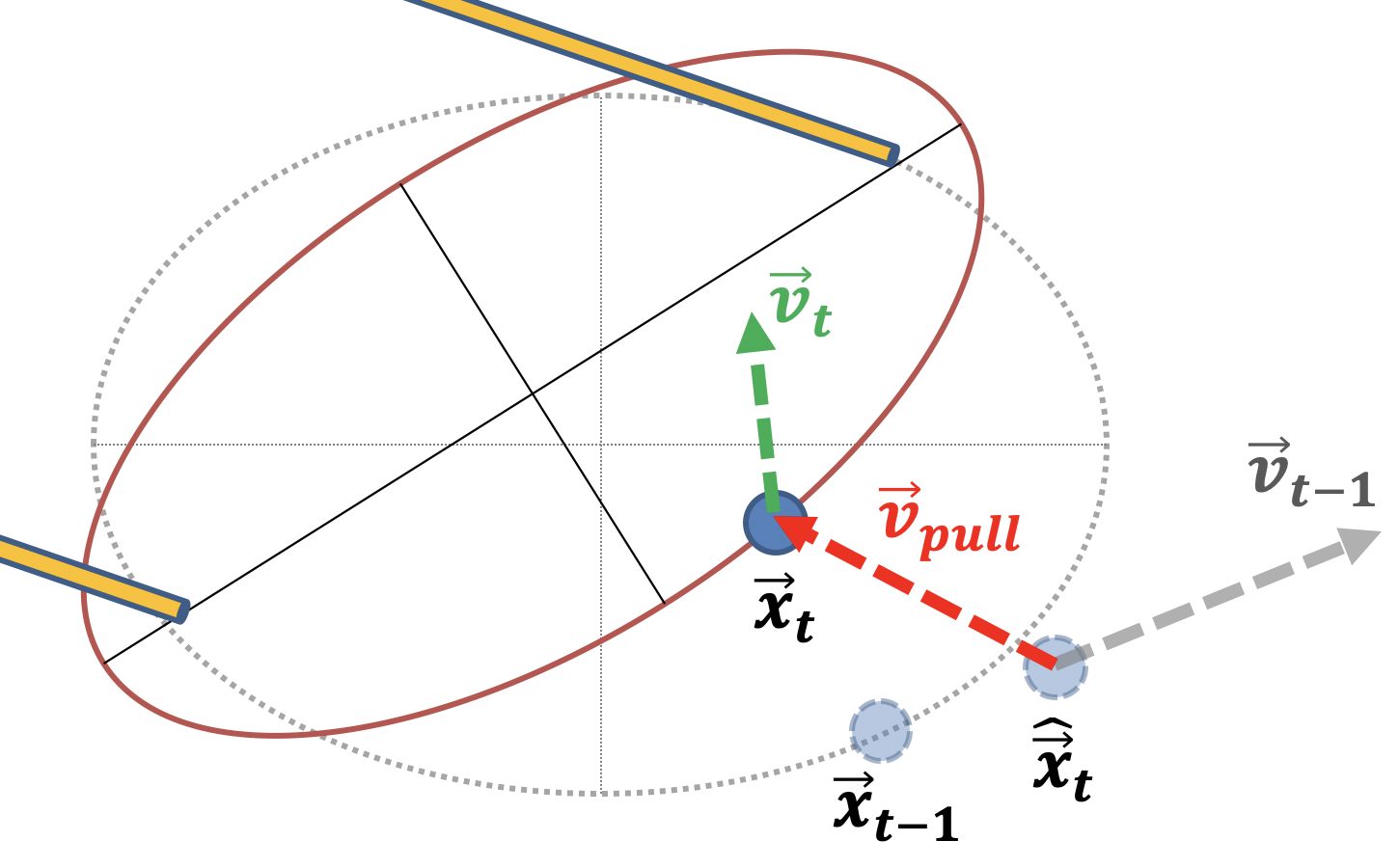}
    \caption{The positions and velocity vectors of the diabolo (in state ON\_STRING) extrapolated from time $t-1$ to $t$, before and after being constrained to the spheroid and adding the pull velocity.}
    \label{fig:ellipse-dynamic}
\end{figure}

\begin{figure}[t]
    \centering
    \includegraphics[width=0.99\columnwidth]{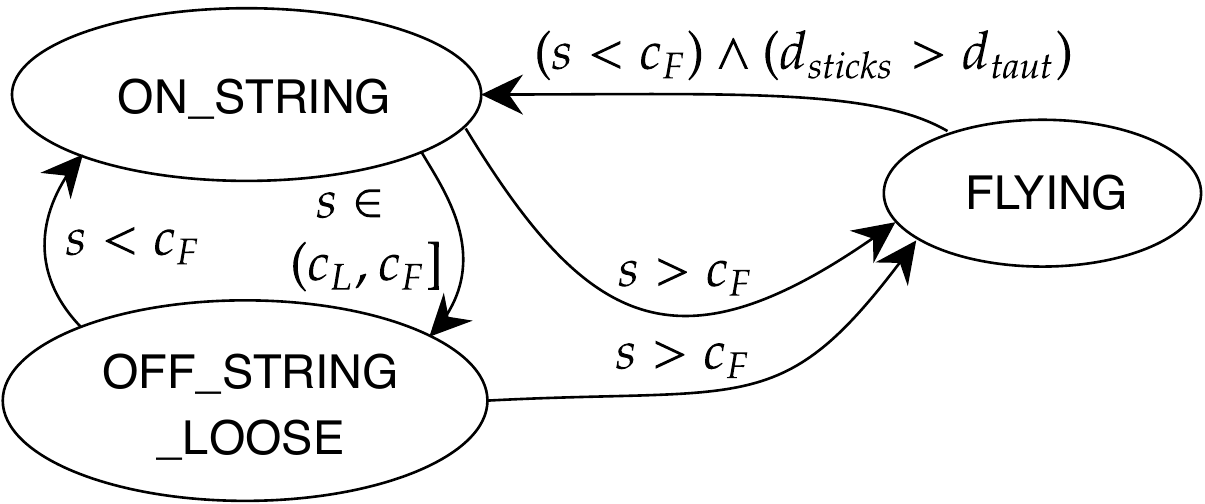}
    \caption{Transition conditions between diabolo states. Position and rotation speed are only affected in state ON\_STRING. Otherwise, the diabolo is moving freely. $s$ is the signed distance of the diabolo's center from the spheroid surface (positive when the diabolo is inside the spheroid). $c_L, c_F$ are critical values (1~cm, 5~cm in our case). 
    OFF\_STRING\_LOOSE represents a state where the string is safely between the diabolo cups, but loose enough not to affect the diabolo's behavior.}
    \label{fig:diabolo-states}
\end{figure}

The diabolo is modeled as a point mass which is affected by the auxiliary spheroid and gravity, and which has three possible states: ON\_STRING, OFF\_STRING\_LOOSE and FLYING.
At each time step, it is updated as follows:

\begin{enumerate}
    \item If the diabolo is on the string, its rotation speed is calculated via the length of string that has passed along the axis, the current rotation speed and a factor representing friction phenomena:
\end{enumerate}

\begin{subequations}
  \label{equation:diabolo-rotation-speed}
  \begin{align}
    d_{r} &= \left | \vec{x}_\mathrm{right}-\vec{x}_\mathrm{diabolo} \right | \\
    \Delta_\mathrm{string} &= (d_{r,t}-d_{r,t-1}) \\
    \omega_{t} &= \omega_{t-1} + \mu\Delta_\mathrm{string}, \mbox{where} \\
    \mu &= \begin{cases}
      \mu_\mathrm{acceleration}, \quad \mbox{when } \Delta_\mathrm{string} > 0 \\
      \mu_\mathrm{deceleration}, \quad \mbox{otherwise}
      \end{cases}
  \end{align}
\end{subequations}

\begin{enumerate}
    \setcounter{enumi}{1}
    \item The diabolo's new position and velocity are extrapolated from the previous time step using the forward Euler method, and the new spheroid is calculated from the sticks' position, as described above. 
    \item The current diabolo state is calculated as a function of the previous status and the current position on the ellipse, as explained in Fig.~\ref{fig:diabolo-states} 
    \item If the diabolo's extrapolated position is outside of the spheroid, it is moved to the closest point on the spheroid instead, and a part of the vector required for this displacement $\vec{v}_\mathrm{pull}$\footnote{This displacement vector is approximated by the normal vector of the spheroid at the new position, as this is easier to obtain.} is added to the diabolo's velocity. 
\end{enumerate}

When the velocity is large or measurements are noisy, this ``pull velocity'' $\vec{v}_\mathrm{pull}$ shown in Fig.~\ref{fig:ellipse-dynamic} can cause unstable behavior, so it is limited by the sum of the movement of the spheroid's origin and the decrease in the spheroid's semi-minor axis (by pulling the strings apart):

\begin{align}
    \vec{v}_\mathrm{pull,capped} = 
    \mathrm{min} (\vec{v}_\mathrm{pull} , \vec{v}_\mathrm{ellipse\_origin} + \vec{v}_\mathrm{ellipse\_edge})
    \label{eq:velocity-damping-parameters}
\end{align}

The pull velocity application can also become unstable when the sticks are moved far apart to launch the diabolo, as the spheroid becomes very ``sharp'' or ``thin'', and the surface normals change abruptly.
To avoid erratic accelerations of the diabolo during throwing, we change the velocity calculation when the distance between the sticks is above a certain threshold (in our system, we chose 5~cm below the 145~cm string length), such that the pull velocity is only applied in the normal direction of the plane defined by the point and the normal vector:

\begin{subequations} \label{equation:cut-spheroid-plane}
  \begin{align}
      \vec{p}_\mathrm{plane} &= (\vec{x}_\mathrm{left} + \vec{x}_\mathrm{right}) / 2 \\
      \vec{n}_\mathrm{plane} &= (1,0,0)^T_\mathrm{world} \times (\vec{x}_\mathrm{left} - \vec{x}_\mathrm{right})
  \end{align}
\end{subequations}

where the world's x-axis points ``forward'' facing away from the robots.
This plane cuts the spheroid in half facing upwards, and keeps the diabolo predictor from making large velocity jumps in the backward/forward direction.

Furthermore, damping factors are applied at each time step, (1) to the added velocity before and after capping, and (2) to the velocity of the diabolo when it is on the string, representing an overall dissipation factor. 
The factors were determined by minimizing the position error when comparing the simulation results with recorded diabolo and stick positions.
The diabolo prediction is done at a time step of one to five ms.

We focused on simulating the diabolo's motion and did not model the effect of forces (excluding gravity) explicitly.
This was motivated by the fact that the dynamic loads would be small compared to our robots' payload, and that we would not use force measurements to estimate the diabolo's state.
We discuss this choice in Sec.~\ref{sec:discussion}. 

\subsection{Stick Motion Generation (Player)}

The diabolo model allows us to predict its motion given a stick trajectory.
We use this to search for a stick trajectory which results in a target diabolo motion, described by a goal state and optional waypoints.

We parametrize the stick trajectory with piece-wise splines that are initialized to describe a motion that showed stable behavior for the target diabolo motion in simulation.
The optimization method is a random walk that optimizes the status vector consisting of the splines' control points and their time.
The residual is the sum of each waypoint's respective costs multiplied by a weight (set for each waypoint and cost).

\begin{itemize}
    \item \textbf{Position:} Euclidean distance
    \item \textbf{Velocity:} The difference in velocity vector magnitude
    \item \textbf{Direction:} The angle between the velocity vectors
\end{itemize}

These costs are calculated between a diabolo goal state (waypoint) and the diabolo at a certain time in its trajectory.
Each waypoint $x_{i}$ is matched to a time $t_{i}$ in the trajectory by finding the state in the trajectory closest to the first waypoint in the whole trajectory.
The same is performed for subsequent waypoints such that $t_\mathrm{start} < t_{i} < t_{i+1} < t_\mathrm{end}$. 

\subsection{Robot Control}

\begin{figure}[t]
    \centering
    \includegraphics[width=0.99\columnwidth]{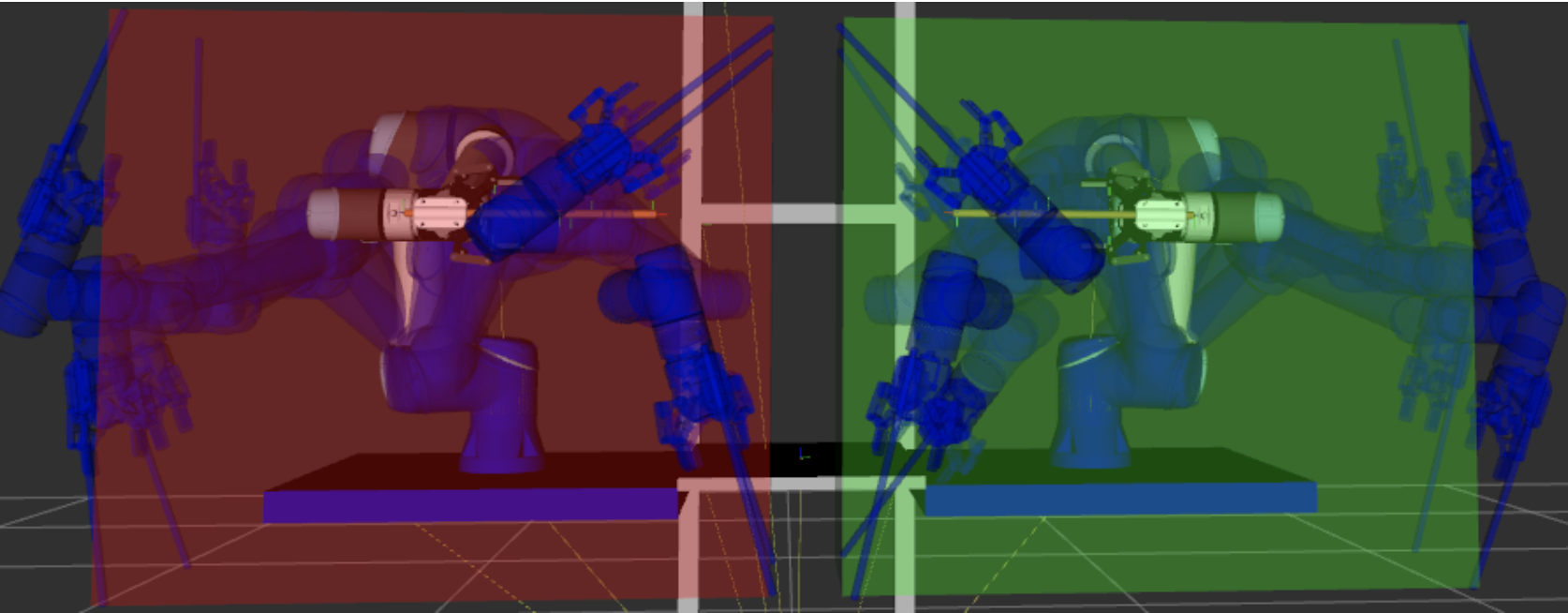}
    \caption{Robot configurations used to interpolate IK seed states inside the workspace. At each blue joint configuration, the stick tip is at a corner of the box which defines the permissible stick positions.}
    \label{fig:ik-interpolation}
\end{figure}

To control the robots, the stick trajectories need to be converted to robot joint trajectories.
Because our Inverse Kinematics (IK) problem is underconstrained, we use a genetic optimization algorithm~\cite{bioik2018} with additional cost functions:
\begin{enumerate}
    \item A cost to keep the first wrist link horizontal
    \item A cost to penalize rotating the last joint
    \item A smaller cost to penalize rotating the second to last joint
\end{enumerate}
Together with the position goal cost applied to the stick tip, this results in an optimization problem with a global minimum, which is almost always solved within five attempts.

Nonetheless, cyclic motions tend to cause the robot to drift, which eventually leads to unfavorable joint configurations which cause failure.
To ensure that this drift does not cause failures, we separate the workspace into boxes with key points at the corners.
For each key point, we calculate and store a robot joint configuration which reaches the point with the stick tip and is smoothly connected to the neighboring key points.
At runtime, we interpolate the joint state based on the neighboring key points' joint configuration to seed the IK solver.
This results in a stable, smoothly connected mapping of natural-looking and safe robot joint configurations.

However, the stick trajectory does not consider the acceleration limits of the robots at runtime.
This is discussed in Section~\ref{sec:discussion}.

The robot control node receives trajectories of about 500--1000~ms length, blends them together and sends the joint states to the robots in a 500~Hz control loop.


\section{Experiments}\label{sec:experiment}

\subsection{Dataset collection}
\label{sec:data-collection}

\begin{figure}[t]
    \centering
    \includegraphics[width=0.99\columnwidth]{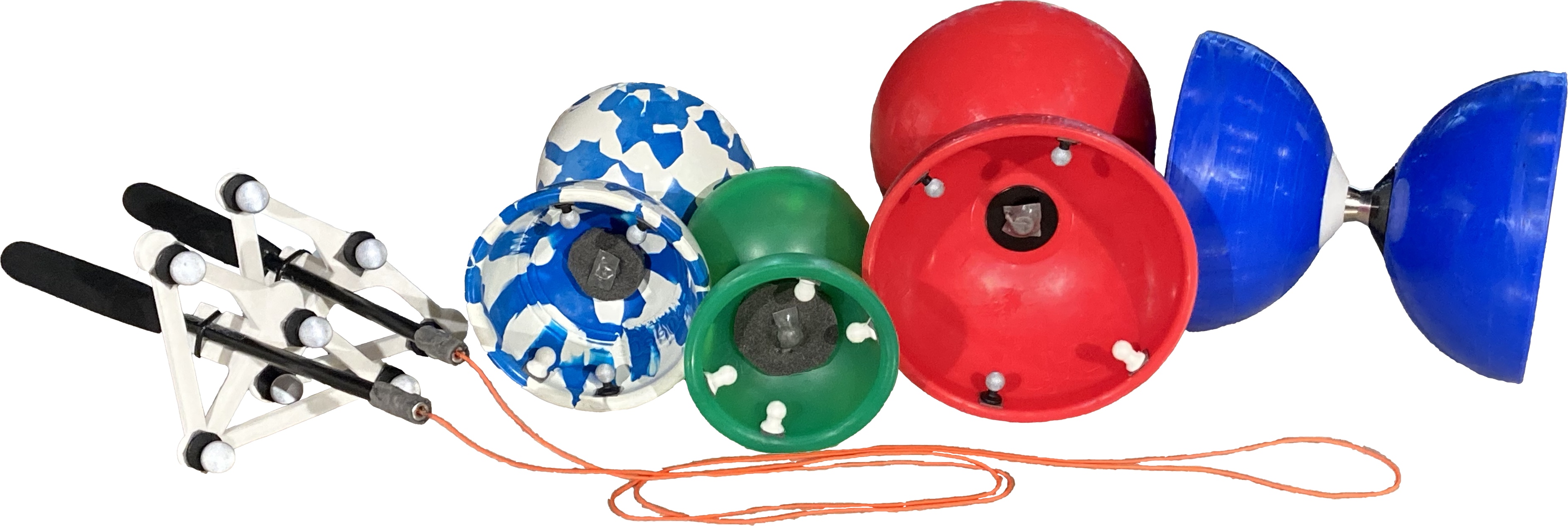}
    \caption{The diabolos and sticks used to record the dataset (Patterned, Green, Red, Blue). Grey markers inside the cups reflect infrared light for Motion Capture, white markers are non-reflective 3D prints. The marker pattern is explained in Sec.~\ref{sec:data-collection}}
    \label{fig:diabolos}
\end{figure}

We recorded about 60~minutes of diabolo play with four diabolos with different characteristics, as shown in Fig.~\ref{fig:diabolos}.
We used three motion capture cameras (FLEX13, OptiTrack) and their software (Motive, OptiTrack) to extract the position and orientation of the sticks and diabolos from reflective infrared markers attached to the sticks and diabolos.
As the diabolos can rotate more than 90 times per second~\cite{SharpeDiaboloAnalysis2008}, we attached the infrared markers on the inside of the cups, so that the centrifugal force presses the markers onto the adhesive tape.

A symmetric pattern of markers is needed to avoid weight imbalances in the diabolo, which would lead to undesirable behavior such as excessive vibration and unstable tilting/yawing.
However, symmetric patterns of reflective markers cannot be resolved uniquely to a rigid body by the motion capture software, so we substituted 3D-printed dummy markers.
The markers do not noticeably affect the tilt/yaw behavior, although they increase the diabolos' rotational inertia.

\begin{table}[t]
    \centering
    \begin{tabular}{l|l|l|l|l}
                       \multicolumn{1}{c|}{\textbf{Color}} & \multicolumn{1}{c|}{\textbf{\begin{tabular}[c]{@{}c@{}}Weight \\ {[}g{]}\end{tabular}}} & \multicolumn{1}{c|}{\textbf{\begin{tabular}[c]{@{}c@{}}Diameter \\ {[}mm{]}\end{tabular}}} & \multicolumn{1}{c|}{\textbf{\begin{tabular}[c]{@{}c@{}}Length \\ {[}mm{]}\end{tabular}}} & \multicolumn{1}{c}{\textbf{\begin{tabular}[c]{@{}c@{}}Axle \\ bearing\end{tabular}}} \\ \hline
    \textbf{Red}       & 284.5   & 127        & 143      & no   \\ \hline
    \textbf{Blue}      & 289.5   & 127        & 143      & yes  \\ \hline
    \textbf{Patterned} & 218.0   & 97         & 114      & no   \\ \hline
    \textbf{Green}     & 141.5   & 82         & 93       & no  
    \end{tabular}
    \caption{Characteristics of the diabolos in the dataset. The eight markers (about 0.6~g each) on each diabolo are not included in the weight.}
    \label{tab:diabolo-facts}
    
\end{table}

\subsection{Neural network predictor}

The previous study~\cite{murooka-diabolo} about robotic diabolo manipulation proposed a learning-based diabolo state predictor for diabolo orientation stabilization control by a humanoid robot.
Following the network implementation of that study, we implemented a learning-based predictor using a deep neural network.
As in the analytical predictor, we assumed that the next diabolo state could be predicted from the current diabolo state, and the next and current stick tip positions.
The network representation $f$ for the predictor is formulated as follows.
\begin{subequations}
  \label{equation:network-f}
  \begin{align}
    \vec{x}_\mathrm{center} =& (\vec{x}_\mathrm{left} - \vec{x}_\mathrm{right})/2 \\
    \vec{x}_\mathrm{diabolo}^{'} =& \vec{x}_\mathrm{diabolo}-\vec{x}_\mathrm{center} \\
    \vec{x}_\mathrm{right}^{'} =& \vec{x}_\mathrm{right}-\vec{x}_\mathrm{center} \\    
    \vec{x}_\mathrm{left}^{'} =& \vec{x}_\mathrm{left}-\vec{x}_\mathrm{center} \\
    \vec{s}_\mathrm{diabolo} =& (\vec{x}_\mathrm{diabolo}^{'}, \vec{v}_\mathrm{diabolo}, \omega_t) \\
    \vec{s}_{\mathrm{diabolo}, t+1} =& f (\vec{s}_{\mathrm{diabolo}, t}, \vec{x}_{\mathrm{right}, t+1}^{'}, \vec{x}_{\mathrm{left}, t+1}^{'}, \nonumber \\
    & \vec{x}_{\mathrm{right}, t}^{'}, \vec{x}_{\mathrm{left}, t}^{'}) 
  \end{align}
\end{subequations}

We translated the diabolo and stick positions to express them relative to the spheroid center, in order to reduce the amount of training data required for the network to generalize.
Expressing the coordinates in world coordinates 

The network consists of three fully-connected layers of 100 units with ReLU activations except for the output layer.
To train the network, we recorded motion capture data of a three-minute diabolo performance with a variety of motions, and extracted the diabolo position and velocity and stick positions.
We implemented the network and training process with PyTorch~\cite{pytorch} and the Adam optimizer~\cite{adam}, respectively.

\subsection{Physical robot setup}

We used two UR5e robot arms mounted at 70~cm height and 110~cm spacing, with wooden diabolo sticks attached via 3D printed holders.

With extended arms and horizontal sticks, the string moved about 5~cm under the load of the blue diabolo with markers, which corresponds to a spring stiffness of approx. 60~N/m.
This elasticity is mainly due to the bending of the sticks, and is almost imperceptible when the sticks are in line with the string and do not experience bending loads.
Our model omits the system's elasticity.


\section{Results}

\subsection{Diabolo model fidelity}

\begin{figure}[t]
    \centering
    \includegraphics[width=0.8\columnwidth]{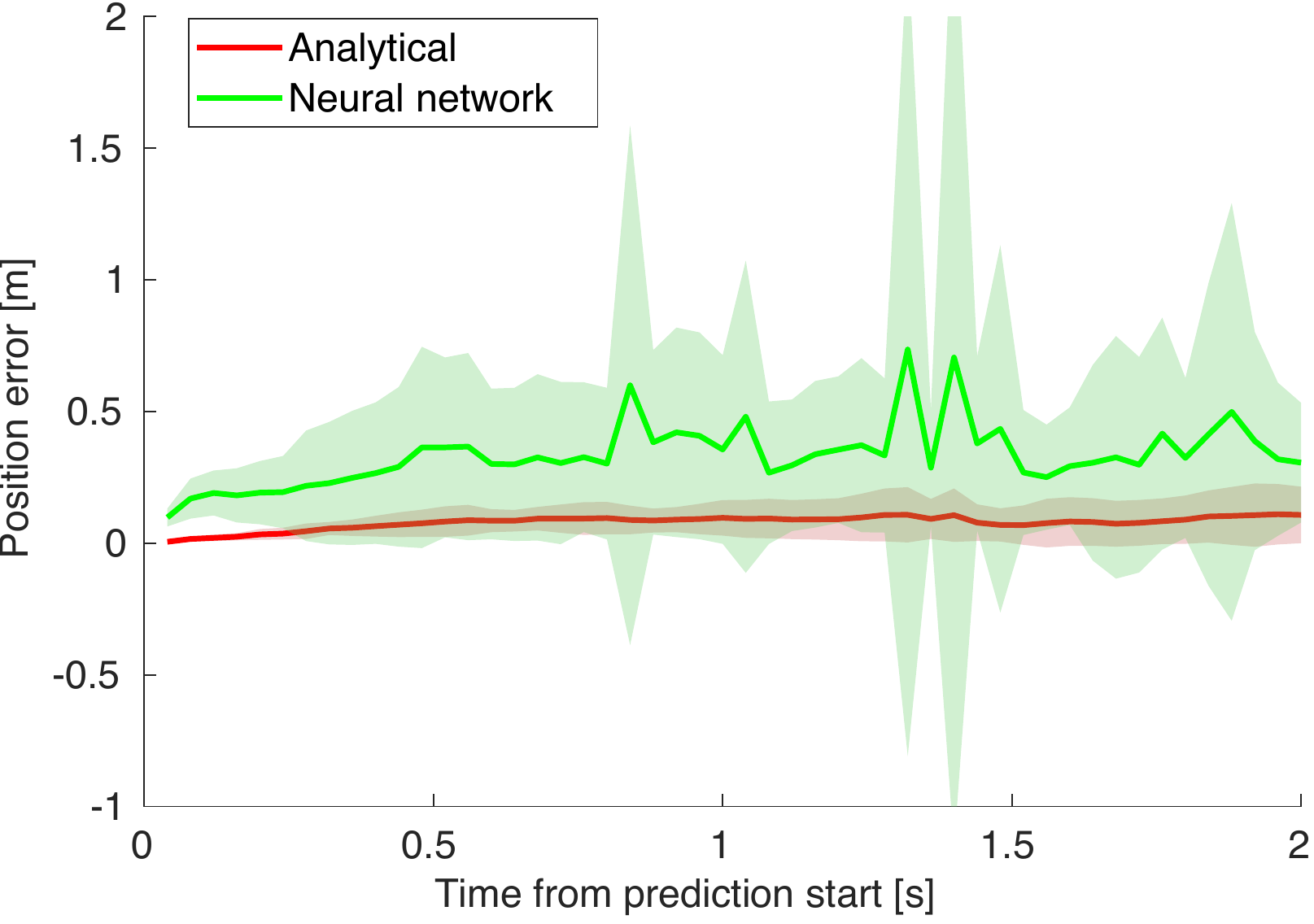}
    \caption{The average evolution of the position error when predicting from different instants in a 40-minute dataset of free diabolo play. 
    The error tends to be larger for dynamic motions.}
    \label{fig:error-evolution}
\end{figure}

To determine the accuracy of our analytical model, we predicted the diabolo's motion given a start state and subsequent stick positions taken from about 40 minutes of recorded data containing a large variety of motions, and then compared the evolution of the position error.
As shown in Fig.~\ref{fig:error-evolution}, the analytical model generally stays close to the real diabolo, whereas the neural network predictor can sometimes diverge significantly.
Tab.~\ref{tab:elementary-motions} shows how predictor performance varies for different kinds of motions.
The neural network predictor diverged for the acceleration motion, causing a very large error.

\begin{figure}[t]
    \centering
    \includegraphics[width=0.8\columnwidth]{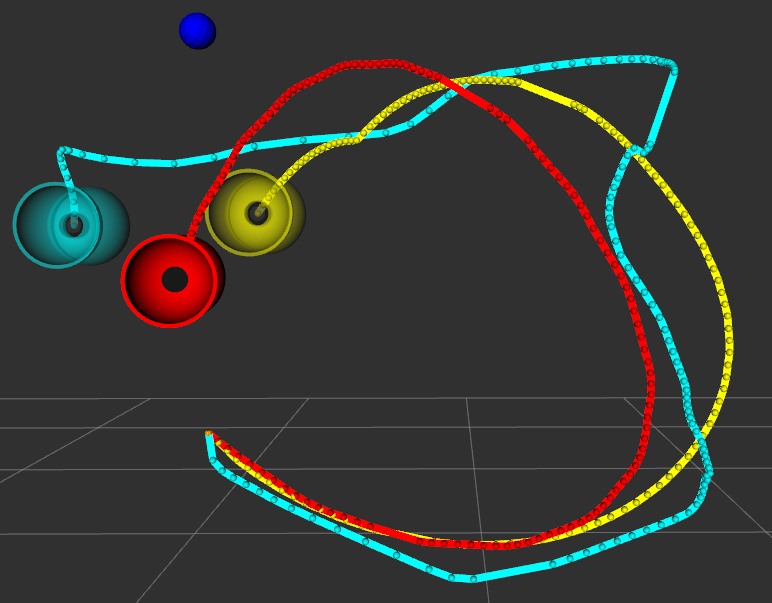}
    \caption{The trajectories of the real diabolo (red), our proposed model (yellow) and the learning-based predictor (cyan). Our analytical model diverges only slightly, while the learning-based method is physically inconsistent.}
    \label{fig:nn-analytical-trajectories}
\end{figure}

Qualitative evaluations of the trajectories as seen in Fig.~\ref{fig:nn-analytical-trajectories} show that the neural network predictor moves in physically inconsistent ways.

\begin{table}[t]
    \centering
    \begin{tabular}{c|c|c}
        \textbf{Motion}& \textbf{\begin{tabular}[c]{@{}c@{}}Avg. error\\ (proposed method)\end{tabular}} & \textbf{\begin{tabular}[c]{@{}c@{}}Avg. error\\ (neural net)\end{tabular}} \\ \hline
        \textbf{Throw} & 0.21 & 0.57 \\ \hline
        \textbf{Hop}   & 0.07    &  0.14     \\ \hline
        \textbf{\begin{tabular}[c]{@{}c@{}}Linear acceleration\end{tabular}}       & 0.03    & 2.81       \\ \hline
        \textbf{\begin{tabular}[c]{@{}c@{}}Circular acceleration\end{tabular}}     & 0.11    & 0.21       \\ \hline
        \textbf{Swinging}     & 0.08    & 0.4 
    \end{tabular}
    \caption{Average position error between predicted and real motion for the proposed method and a learning-based classifier.}
    \label{tab:elementary-motions}
\end{table}

\subsection{Motion generation}

We performed linear acceleration and circular acceleration motions in real-time simulation to evaluate the feasibility of our proposed motion generation and control method.
We implemented the system shown in Fig.~\ref{fig:system-layout} in ROS Melodic, C++ and Python, and used Gazebo 9.0 to simulate the bottom two elements (obtaining the diabolo state directly from simulation), and were able to perform the circular acceleration motion stably by setting four sequential goals in a circle as shown in Fig.~\ref{fig:intro}.

\subsection{Robot performance}

By setting a diabolo goal state with a velocity vector pointing upwards, we were able to generate several acceleration and throwing motions.
However, due to the high numerical sensitivity for this motion and the unmodeled elasticity in the system, the jump height of the diabolo in the real world differed.
During the cyclic acceleration movements, the diabolo also tended to fall ``out of phase''.
We conclude that timing has a significant effect on the motions' performance and effect, but that the overall dynamics of the diabolo are sufficiently close to reality.


\section{Discussion}\label{sec:discussion}

The results clearly show that our analytical model outperforms the learning-based predictor by a significant margin. 
This network is structured according to the previous study~\cite{murooka-diabolo} to predict the next state from the current state and action. 
Although the previous study succeeded in diabolo orientation stabilization control, the result in this application is significantly worse than our proposed analytical model, up to occasional divergence of the predicted trajectory.
This is because the previous study assumed that the diabolo position would not be affected by the stick position, and that there is no agile motion of sticks dealt with in this study.
Moreover, they only use the diabolo orientation for the state and stick position, which is projected to a two-dimensional space for the action.
This simplification leads to a narrowed action and state space.
In contrast, this study deals with agile motions where the diabolo position moves in a wider space, which makes the training of the network significantly more sensitive to the dataset.

Recent studies of modeling for robotic manipulation utilize both analytical modeling and learning-based methods.
They first build the analytical model of the dynamics physically, then construct a network to approximate the residual between the analytical and actual model as residual physics~\cite{tossingbot}.
This combined method of the analytical and learning-based modeling is seen in other studies~\cite{mcube-residual}~\cite{tunenet}.
The analytical model we propose in this work can also be used for residual physics learning, which we will consider in future work.

Our proposed system is also able to generate motions that follow diabolo goal states.
A limitation is that we perform the stick trajectory generation without knowledge of the robots, which means that the robot trajectories' speed and accelerations may exceed the robots' capabilities.
We work around this by using robots with sufficiently high velocity and acceleration limits.

Our diabolo model is simplified in a number of ways, which we plan to extend in future work:

\begin{enumerate}
    \item The diabolo's tilt and yaw are how the instability of the diabolo manifests, and maintaining a stable diabolo orientation is one of the main challenges for beginner players. This is the main focus of our future work.
    \item Wrapping the string around the diabolo's axis (or looping a diabolo around the string) modifies the behavior of the diabolo and is required for certain manipulation techniques.
    \item The forces acting on the sticks could be used to estimate the diabolo's state, just as skilled human players rely on their proprioception more than their vision.
\end{enumerate}


\section{Conclusion}\label{sec:conclusion}

We proposed a diabolo model that can be used for robot learning, and a system to generate motion sequences for and control a robot system playing diabolo.
The results showed that our model significantly outperforms a purely learning-based predictor, and that the system can generate stable motions that accelerate the diabolo.
The model as well as the dataset we recorded will be released open-source~\footnote{Before publication of this paper}, so that the community can test learning algorithms on this new task.
We hope that its unique set of characteristics and incremental challenges will facilitate the development of new approaches and techniques.
Future work will extend the diabolo model to include gyroscope dynamics and string wrapping, testing residual physics learning to improve the model, and training machine learning agents to perform tricks.



\bibliographystyle{IEEEtran}
\bibliography{refs.bib}

\end{document}